# Speech Signal Filters based on Soft Computing Techniques: A Comparison

Sachin Lakra
Dept. of Information Technology
Manav Rachna College of Engg
Faridabad, Haryana, India
and R. S., K L University
sachinlakra@yahoo.co.in

T. V. Prasad
Dean, Industrial Consultancy
& Associate Dean, Academics
Lingaya's University
Faridabad, Haryana, India
tvprasad2002@yahoo.com

G. Rama Krishna
Dept. of Comp. Sc. & Engg.
K L University
Vijayawada,
Andhra Pradesh, India
ramakrishna_10@yahoo.com

*Abstract*—The paper presents a comparison of various soft computing techniques used for filtering and enhancing speech signals. The three major techniques that fall under soft computing are neural networks, fuzzy systems and genetic algorithms. Other hybrid techniques such as neuro-fuzzy systems are also available. In general, soft computing techniques have been experimentally observed to give far superior performance as compared to non-soft computing techniques in terms of robustness and accuracy.

*Keywords- Speech Signal Filtering, Soft Computing, Multi Layer Perceptron Filter, Genetic Time Warping Filter, Adaptive Neuro-Fuzzy Filter, Adaptive Recurrent Neuro-Fuzzy Filter.*

## I. INTRODUCTION

Speech Signal filtering is an active research area in speech processing and soft computing techniques are now being employed for the process. Various approaches have been used in the past for filtering speech signals. One approach to filter noise is a linear filter called a band pass filter which is unsuitable for filtering speech signals since the number of possible frequencies in the human audible range at which audio signals occur in the real world is very large. Besides this, a band pass filter cannot handle fuzzy rules and fuzzy values representing ranges of frequencies along with not being able to handle them in a robust manner by handling imprecision and time variance. More robust, more effective and more efficient techniques from the realm of soft computing are being applied to solve fundamental problems. Some instances of such application include co-active neuro-fuzzy inference systems for the XOR problem [11], fuzzy mathematics for paralinguistic content elimination from a speech signal [10] and hybrid techniques for speech signal filtering.

## II. SOFT COMPUTING TECHNIQUES USED IN FILTERING OF SPEECH SIGNALS

Soft Computing techniques that have been used for filtering speech signals include Neural Networks, Genetic Algorithms, Fuzzy Systems and Hybrid techniques such as Neuro-Fuzzy Systems. Each technique improves upon already existing non-soft computing techniques including Elman Filter, Dynamic Time Warping and other linear filters. The filters that have been compared in this paper include the Multi-Layer Perceptron Filter [8] based on Neural Networks, the Genetic Time Warping Filter [3] based on Genetic Algorithms and two models, namely, the Adaptive Neuro-Fuzzy Filter [1] and the Adaptive Recurrent Neuro-Fuzzy Filter [13], based on the hybrid neuro-fuzzy technique.

Fig. 1 to Fig. 4 show the structures of each of the four filters. Fig. 5 to Fig. 9 depict the performance of the filters in comparison to various linear filters. Table 1 compares the four techniques with each other in terms of various parameters.

## III. SOME SPEECH SIGNAL FILTERS BASED ON SOFT COMPUTING TECHNIQUES

### A. Multi Layer Perceptron Filter

An FIR digital filter has been used as in [8] to train a neural network. The experimental results in Fig. 5 and Fig. 6 show that using neural networks in filtering noise from a speech signal is a more robust and powerful technique than other traditional algorithms. Furthermore, the FIR digital filter used gives a fast convergence and provides results close to the global optimal. A Multi Layer Perceptron (MLP) was trained with different training algorithms and compared with an FIR digital filter in terms of its performance and computational complexity. It was found that the training algorithm selected to train the neural network being used as a filter is vital and significantly affects the final results [8].

### B. Genetic Time Warping Filter

The technique of dynamic time warping (DTW) is commonly used to assess the similarity of two different speech utterances. In real-time applications, there are some limitations on the use of DTW:

(1) the exact endpoint identification of utterances and
(2) the use of a constant normalization factor instead of using a real-time factor.

The first limitation leads to the problem of low robustness in speech recognition whereas the second relates to the accuracy of the algorithm used. In view of these, a GA-based time-warping algorithm (GTW) proposed in [9] has been used to improve the global searching ability of DTW to resolve the two problems. There are two stages in this technique, namely, time-warping path identification followed by application of a genetic algorithm on the identified paths (Fig. 2) to elicit the best among them. In the latter stage, every chromosome, each representing a time-warping path, has its own length, which is different from the conventional fixed-length chromosome.

Since the warping paths are directly stored as chromosomes, this leads to the possibility of n-best warping path solutions being obtained without extra computation time, although the solutions may not necessarily be optimum [3].

## C. Fuzzy Filters

The fuzzy part of the fuzzy filters allows filtering out of noise components as these filters allow only those components of the speech signal to pass through which are not part of the noise by considering inputs as falling into fuzzy sets which can be mapped to outputs by fuzzy rules. The concepts of fuzzy theory must usually be combined with another soft computing technique to give a hybrid technique for best results. Two such hybrid filters are the Adaptive Neuro-Fuzzy Filter and the Adaptive Recurrent Neuro-Fuzzy Filter.

*1) Adaptive Neuro-Fuzzy Filter:* An approach based on the hybrid technique of fuzzy theory combined with a neural network is an Adaptive Neuro-Fuzzy Filter (ANFF) [1]. The ANFF is essentially a feedforward multilayered connectionist network which can learn on its own according to either numerical training data or linguistic expert knowledge represented by fuzzy rules. The adaptive part in such a filter includes the construction of fuzzy rules (structure learning), and the tuning of the parameters of membership functions (parameter learning).

In the structure learning phase, fuzzy if-then rules have been identified as in [1] on the basis of the matching of input-output clusters. In the parameter learning phase, an adaptation algorithm similar to backpropagation has been developed as in [1] to minimize the output error. Initially, there are no hidden nodes, i.e., no membership functions or fuzzy rules, and both the structure learning and parameter learning phases are executed concurrently as the adaptation proceeds. However, if some linguistic information about the design of the filter is available, such knowledge can be put into the ANFF to form an initial structure with hidden nodes.

Two major advantages of the ANFF are: 1) a priori knowledge can be incorporated into the ANFF which allows the fusion of numerical data and linguistic information in the filter; and 2) the number of hidden nodes, need not be given in a predetermined manner, since the ANFF can automatically find its optimal structure and parameters. In contrast to traditional fuzzy systems where the input-output spaces are partitioned in a manner structurally similar to a grid leading to a combinatorial growth of fuzzy rules as the input-output dimensions increase, irregular partitions are done in the ANFF according to the distribution of training data resulting in fewer fuzzy rules being generated. In the ANFF, available expert knowledge is used as a set of initial fuzzy rules. Carpenter's Fuzzy Adaptive Resonance Theory for Clustering by space partitioning or grid partitioning is used for structure learning, whereas for the adaptation of membership functions, the backpropagation algorithm is used to implement the adaptation as part of parameter learning. Fuzzy logic rules are then derived as a part of the adaptive process. The structure of the ANFF is shown in Fig. 3 whereas its performance is compared with linear filters of various orders in Fig. 8.

*2) Adaptive Recurrent Neuro-Fuzzy Filter:* Another novel approach for filtering noisy speech signals along with enhancing them is the adaptive recurrent neuro-fuzzy filter (ARNFF) [13]. The speech enhancement scheme consists of using two microphones that receive a primary and a reference input source respectively, and the proposed ARNFF attenuates the noise distorting the uncorrupted speech signal in the primary channel. The ARNFF is a connectionist network that can he translated effortlessly into both a set of dynamic fuzzy rules and state-space equations. An effective learning algorithm, consisting of a clustering algorithm for the structure learning and a recurrent learning algorithm for the parameter learning, have been adopted for the ARNNF construction as in [13]. The basic concept used in the ARNFF is called Active Noise Control (ANC), which uses the principle of destructive interference to generate an antinoise signal to cancel the noise distorting the speech signal. The ANC allows improvements in suppressing acoustic noise at low frequencies with benefits in much smaller size, weight, volume as well as cost [7,12].

Traditionally, ANC has been implemented by using linear filters such as the FIR and the IIR. Recently, nonlinear filters including Volterra filters [6] and neural-network-based filters [5] have been proposed to provide alternatives for the problems posed by linear filters. However, the order of the input variables of these filters, which significantly affects the size of the networks as well as the overall filtering performance, has to be determined in advance leading to non-existence of adaptiveness. Moreover, serious performance degradation has been identified in case of the existence of longer delays in the processing environment [2].

To solve these problems, the incorporation of fuzzy basis functions, possessing universal approximation capability [4], with dynamic elements, into a neural network structure to form a recurrent neuro-fuzzy filter has been done to develop the ARNFF as in [13]. The structure learning algorithm used not only automates the construction of the proposed filter but identifies a thrifty filter structure as well. Since the dynamic elements can memorize the past input history, the proposed filter only requires two inputs, i.e., the current variable and one time-lagged input variable.

The use of dynamic elements enables the proposed ARNFF architecture to extract state-space equations from its internal structure. Moreover, the interpretation and extraction of dynamic fuzzy IF-THEN rules from the resulting ARNFF architecture is still transparent and without much complexity although the structure of the ARNFF as a whole is rather complex. To implement the above idea, a fuzzy-basis-function (FBF) network has been employed in [4] as the antecedent part of a conventional (static) fuzzy system. The first four layers comprise a Fuzzy Basis Function Network. Layer 5 implements a recursive recurrent learning algorithm. Layer 6 is the output layer. The link weights in this layer represent the singleton constituents of the corresponding output variables. The output nodes integrate all the "states" from Layer 5 with the corresponding singleton constituents

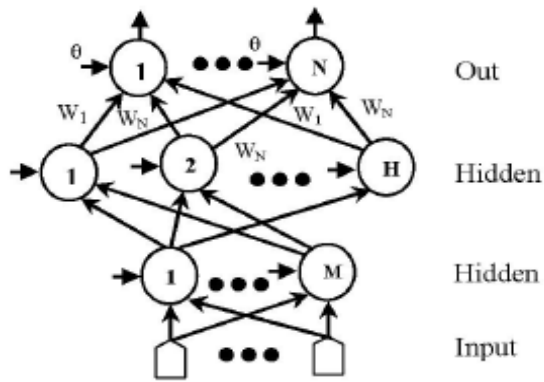

Figure 1. The Structure of the MLP Filter [8]

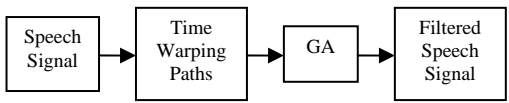

Figure 2. The Structure of the GTW Filter [9]

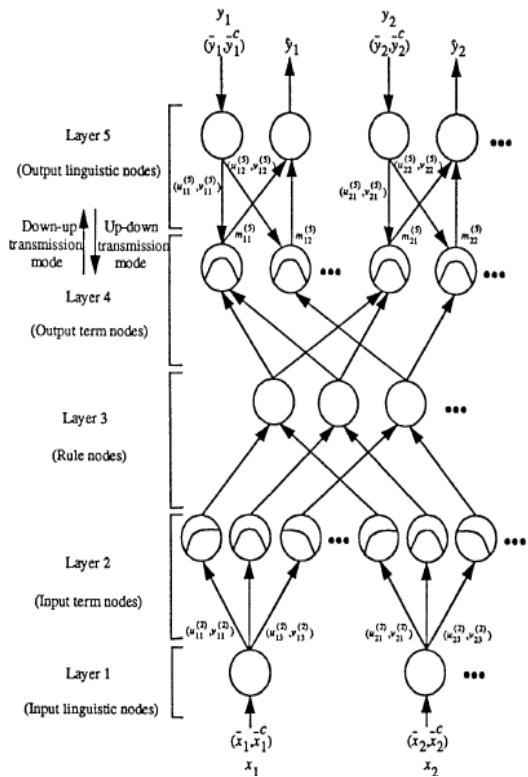

Figure 3. The Structure of the ANFF Filter [1]

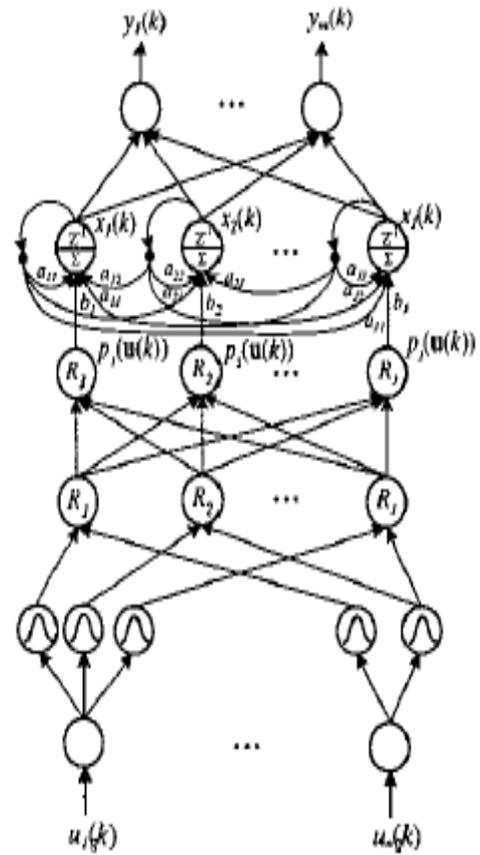

Figure 4. The Structure of the ARNFF Filter [13]

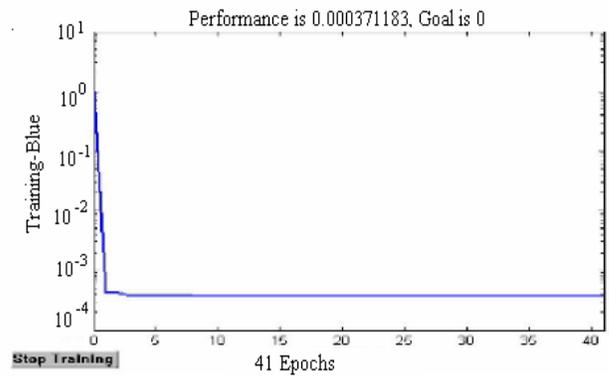

Figure 5. The trained MLP Filter [8]

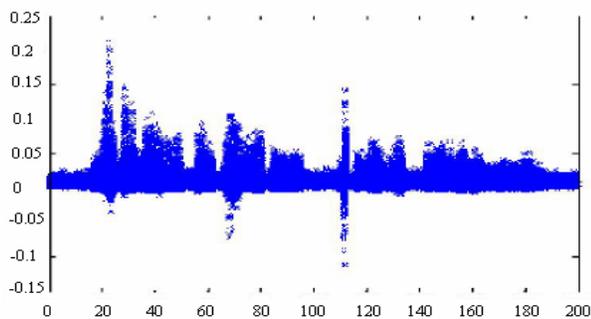

Figure 6. Response for the output of the trained MLP Filter[8]

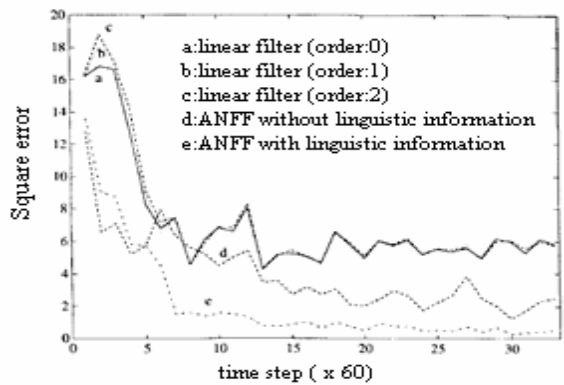

Figure 8. The Performance of the ANFF Filter [1]

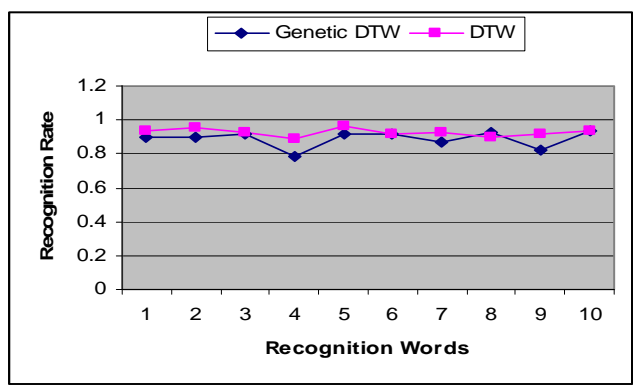

Figure 7. The Performance of the GTW Filter [3]

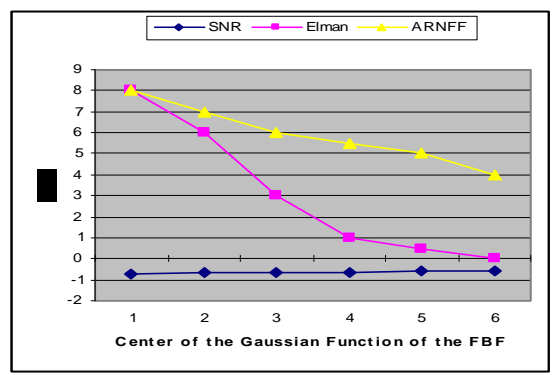

Figure 9. The Performance of the ARNFF Filter [13]

and act as a defuzzifier. The structure of the ARNFF is depicted in Fig. 4 and its performance is compared with the Elman filter in Fig. 9.

## IV. CONCLUSION

Thus, the use of non-linear soft computing techniques provides a more robust, more effective and more efficient method to filter speech signals. Each technique applied individually improves the results by a certain margin and a hybrid technique combining either of the techniques provides the best method among the soft computing methods.

TABLE I. COMPARISON OF THE FOUR MODELS FOR SPEECH SIGNAL FILTERING BASED ON SOFT COMPUTING TECHNIQUES

| Parameter | MLP Filter | GTW Filter | ANFF Filter | ARNFF Filter |
|---|---|---|---|---|
| Data Handled | Speech Signal | Speech Signal | Speech Signal | Speech Signal |
| Structural Complexity | Medium | Medium | High | High |
| Types of noise handled | Gaussian Noise | All types of noise as a GA is a universal optimizer. | White Noise | Long delay noisy speech |
| Accuracy | Better than linear filters | Upto 10% better than DTW | Upto 20% better than Linear filter | Upto 50% better than Elman filter |
| Types of inputs | Crisp | Crisp | Crisp or Fuzzy | Crisp or Fuzzy |
| No. of stages/ layers | Layers:4 | Stages:2 | Layers:5 | Layers:6 |
| Names of stages/layers | Layer 1:Input Layer Layer 2:Hidden Layer 1 Layer 3:Hidden Layer 2 Layer 4:Output | Stage 1:Time Warping Stage Stage 2:Application of GA on a set of Time-Warping paths | Layer 1:Input Linguistic Nodes Layer 2:Input Term nodes (Structure Learning) Layer 3:Rule nodes(Parameter Learning) Layer 4:Output term nodes Layer 5:Output Linguistic nodes | Layers 1 to 4 form a Fuzzy Basis Function Network Layer 1:Input Linguistic Nodes Layer 2:Input Term nodes (Structure Learning) Layer 3:Rule nodes (Parameter Learning) Layer 4:Normalization Layer 5:Dynamic Feedback Layer 6:Output Linguistic nodes |
| Algorithms used at each stage | Layer 1:NA Layer 2:Backpropagation Layer 3:-Backpropagation Layer 4:NA | Stage 1:Time Warping Stage 2:Genetic Algorithm | Layer 1:NA Layer 2:Clustering by Grid Partitioning or Space Partitioning based on Carpenter's Fuzzy Adaptive Resonance Theory Layer 3:Backpropagation Algorithm Layer 4:NA Layer 5:NA | Layer 1:NA Layer 2:Clustering by Grid Partitioning or Space Partitioning based on Carpenter's Fuzzy Adaptive Resonance Theory Layer 3:Recursive Recurrent Learning Algorithm Layer 4:MCA Clustering Algorithm Layer 5:Recursive Recurrent Learning Algorithm Layer 6:NA |
| Types of output | Crisp | Crisp | Crisp or Fuzzy or both | Crisp or Fuzzy or both |
| Advantages | More robust than traditional techniques | More accurate than DTW | 1.A priori knowledge can be incorporated into the ANFF which makes the fusion of numerical data and linguistic information in the filter possible 2.No need of predetermined number of hidden nodes 3.Can find its optimal structure and parameters automatically 4.Greater Flexibility due to Dynamic Partitioning of Input and Output Spaces. 5.Can handle linguistic information as well as non-linguistic information 6.Can generate fuzzy rules as it is adaptive | 1.A more compact filter structure 2.No a priori knowledge needed for the exact lagged order of the input variables 3.A better performance in long-delay environment. |
| Disadvantages | Less accurate than other filters based on soft computing techniques | 1.Requires 10% more computation time than DTW. 2.The exact output which can be obtained is unpredictable. 3.Not useful in real-time applications. | Uses some initial knowledge in the form of fuzzy rules. | Too complex |
| Adaptability | No | No | Yes | Yes |
| Learning | Yes | No | Yes | Yes |